\documentclass[journal]{IEEEtran}
\usepackage{ifpdf}
\ifCLASSINFOpdf
  \usepackage[pdftex]{graphicx}
  \usepackage{graphbox}
  \graphicspath{{figures/}}
  \DeclareGraphicsExtensions{.pdf,.jpeg,.png}
\else
  \usepackage[dvips]{graphicx}
  \graphicspath{{figures/}}
  \DeclareGraphicsExtensions{.eps}
\fi
\usepackage{amsmath}
\usepackage{amssymb}
\usepackage[bookmarks=false]{hyperref} 
\usepackage{colortbl}
\usepackage{booktabs}
\usepackage[per-mode=symbol,range-phrase=--,range-units=single]{siunitx}
\usepackage[table,dvipsnames]{xcolor}
\usepackage{cite}
\usepackage{nicefrac}
\usepackage{multirow}
\usepackage{calc}
\usepackage{printlen}
\usepackage{algorithm}
\usepackage{listings}
\usepackage{bbding}
\usepackage{tikz}

\DeclareSIUnit\year{yr}
\DeclareSIUnit\week{weeks}

\newif\iffinal
\finalfalse 

\iffinal
  
  \newcommand{\todo}[1]{}
  \newcommand{\ts}[1]{}
\else
  
  \newcommand{\ts}[1]{\textcolor{blue}{\textbullet\ \texttt{#1}}\\}
  \newcommand{\todo}[1]{\textcolor{red}{\textbf{TODO} {#1}}}
\fi

\begin{document}
\title{PixelDINO: Semi-Supervised Semantic Segmentation for Detecting Permafrost Disturbances}

\author{Konrad~Heidler,~\IEEEmembership{Student~Member,~IEEE},
        Ingmar~Nitze,
        Guido~Grosse,
        and~Xiao~Xiang~Zhu,~\IEEEmembership{Fellow,~IEEE}%
\thanks{%
  K.~Heidler and X.~Zhu are with the
  Chair of Data Science in Earth Observation (SiPEO),
  Department of Aerospace and Geodesy,
  School of Engineering and Design,
  Technical University of Munich (TUM),
  Munich, Germany.
  E-mails: k.heidler@tum.de; xiaoxiang.zhu@tum.de
}
\thanks{
  I.~Nitze and G.~Grosse are with the
  Permafrost Research Section,
  Alfred Wegener Institute Helholtz Centre for Polar and Marine Research,
  Potsdam, Germany.
  E-mails: ingmar.nitze@awi.de, guido.grosse@awi.de
}%
\thanks{
    G.~Grosse is also with the
    Institute of Geosciences, University of Potsdam, Potsdam, Germany.
}
\thanks{
This work was supported by the
BMWK project ML4Earth.
XZ was further supported by the BMBF future lab AI4EO.
GG and IN were further supported by the projects HGF AI-CORE, and NSF Permafrost Discovery Gateway.}
}
\markboth{SUBMITTED TO IEEE TRANSACTIONS ON GEOSCIENCE AND REMOTE SENSING}%
{Heidler \MakeLowercase{\textit{et al.}}: TODO}

\maketitle
\begin{abstract}
  Arctic Permafrost is facing significant changes
  due to global climate change.
  As these regions are largely inaccessible,
  remote sensing plays a crucial rule in better understanding
  the underlying processes not just on a local scale, but across the Arctic.
  In this study, we focus on the remote detection of retrogressive thaw slumps (RTS),
  a permafrost disturbance comparable to landslides induced by thawing.
  For such analyses from space, deep learning has become an
  indispensable tool,
  but limited labelled training data remains a challenge for
  training accurate models.
  To improve model generalization across the Arctic
  without the need for additional labelled data,
  we present a semi-supervised learning approach
  to train semantic segmentation models to detect RTS.
  Our framework called PixelDINO is trained in parallel
  on labelled data as well as unlabelled data.
  For the unlabelled data,
  the model segments the imagery into self-taught pseudo-classes
  and the training procedure ensures consistency of these pseudo-classes
  across strong augmentations of the input data.
  Our experimental results demonstrate that PixelDINO can improve model performance
  both over supervised baseline methods as well as existing semi-supervised
  semantic segmentation approaches,
  highlighting its potential for training robust models that generalize
  well to regions that were not included in the training data.
  The project page containing code and other materials for this study can be found at \url{https://khdlr.github.io/PixelDINO/}.
\end{abstract}

\IEEEpeerreviewmaketitle

\section{Introduction}

\IEEEPARstart{I}{n} step with global climate change, permafrost is changing rapidly.
Rising temperatures in the Arctic have large implications for
perennially frozen soil
which can destabilize upon thawing of ice-rich ground.
Not only does ice-rich permafrost thaw pose risks to local infrastructure
like roads, pipelines or buildings~\cite{hjort2022_impacts},
but it is also tightly coupled to the global climate system through the
permafrost carbon feedback ~\cite{schuur2015_climate}.
Due to these intricate connections,
permafrost temperature is one of the Essential Climate Variables defined by the
Global Climate Observing System~\cite{gcos2016_global}.

More than a tenth of Earth's land surface is underlain by permafrost~\cite{obu2021_how}.
Owing to their remoteness and sparse population,
these areas are often difficult to access physically.
Therefore, in-situ measurements are only available for specific study sites
at specific dates when expeditions visited that site
or when data is collected through local sensors.
Therefore, remote sensing techniques are a valuable complementary method
that provides a cost-effective way for spatially consistent monitoring of permafrost regions,
and a useful approach for upscaling and understanding of broad spatio-temporal dynamics of permafrost thaw processes.
To further improve the efficiency of remote sensing monitoring for these applications,
machine learning techniques offer great potential in automating tedious annotation tasks.

Following the definition of permafrost
as ground at or below 0 degrees Celsius for at least two consecutive years,
permafrost is generally a subsurface phenomenon.
This makes it infeasible to directly observe permafrost
from satellite measurements in nearly all cases.
Instead, remote sensing studies usually
focus on monitoring surface proxies that are highly correlated with
either the presence or degradation of permafrost~\cite{bartsch2023_permafrost}.
For this study, we will focus on so-called retrogressive thaw slumps (RTS).

RTS are a specific type of landslide that occurs when ice-rich parts of permafrost thaw.
As the ground destabilizes due to the thawing,
soil collapses along slopes and shores,
leading to landslide-like terrain deformations.
Generally, RTS are rather small features with individual RTS usually
measuring less than \SI{10}{\hectare} in area.
The largest RTS, so-called megaslumps, in Northwestern Canada and Siberia exceed \SI{40}{\hectare}.
While they vary in appearance and dynamics,
they form due to specific local environmental conditions like slope,
landscape history, ground temperature, and disturbances.
They typically occur in glacial moraines with preserved remnant glacial ice,
syngenetic ice-rich yedoma permafrost, or marine deposits, which were raised due to isostatic uplift.
Understanding and quantifying RTS dynamics is important as they pose potetnial hazards to infrastructure, directly affect water quality in downstream aquatic environments, and locally mobilize large amounts
of formerly frozen sediment and organic matter.~\cite{nitze2018_remote}

Other than permafrost itself,
permafrost degradation landforms like RTS are indeed possible to detect in optical satellite
imagery due to their distinct shape and spectral signature
compared to the surrounding regions.
This makes them a viable target of study via remote sensing methods.
At the same time,
the detection of such features in satellite imagery is not without challenges.
Retrogressive thaw slumps in permafrost regions are often
hard to detect due to their widespread distribution, small size, and their varying stages of activity.
Further, optical remote sensing is inhibited by snow cover,
frequent cloud cover, and polar night for large parts of the year,
so that features can only be reliably detected during the summer months.

Machine learning, specifically deep learning,
can automate the identification of retrogressive thaw slumps from satellite imagery.
Existing studies often achieve mixed results,
which in many cases can be attributed to the
algorithms' requirements for an extensive collection of
labelled training data that is hard to acquire in large volumes~\cite{nitze2021_developing}.
While decent prediction results are obtained for selected study sites,
accurate pan-Arctic generalization remains an elusive goal.

This is symptomatic of a much larger pattern.
In many remote sensing tasks,
the area of interest is quite large or even global,
but only limited training data for a few sites is available,
and even within those sites, the actual targets can remain sparsely distributed.
This issue is particularly present for segmentation tasks
where the labelling of individual pixels is especially costly.
The segmentation of RTS in permafrost regions is a
prime example for this issue.
The landscapes affected by RTS are extremely diverse and cover several
climate zones, ecosystems and landscape types.
While increasing the available training data through additional
labelling efforts is always an option,
it comes at a large labor cost for the involved domain experts.
Therefore, this study explores how to make models better generalize
to new locations in such settings.

After the models are trained,
they will then be applied to predict targets for a larger region of interest.
In order to keep the computational need for this inference stage manageable,
an additional requirement for our framework is
to avoid increasing the computational resources needed for model inference.
Therefore, we look at strategies that can be applied already at training
time to make a model of the same architecture
to generalize better and be more robust in its predictions.

In an attempt to tackle this issue from a methodological angle, 
we explore semi-supervised learning
for improving model performance without the need
for additional training data.
This is a strategy that can be applied at training time
to encourage better model generalization.
It allows for the inclusion of unlabelled satellite
imagery into the training process.
While labelling satellite imagery is a laborious task,
the underlying satellite imagery is openly available,
meaning that unlabelled imagery over permafrost regions
is readily available.
Therefore, semi-supervised learning methods are exceptionally well-suited 
for these remote sensing tasks.

In this study, we propose a new framework for semi-supervised semantic
segmentation called \emph{PixelDINO}.
Our framework builds on the successful self-supervised learning framework DINO~\cite{caron2021_emerging},
which was originally developed to learn features for image classification.
We adopt the main idea behind DINO,
namely self-distillation,
to pixel-wise prediction tasks like semantic segmentation
and then combine it with a regular supervised learning procedure
into a semi-supervised learning framework.
Finally, we present experimental results for the task of RTS detection,
where we demonstrate that PixelDINO outperforms both supervised baseline methods
and other semi-supervised semantic segmentation approaches.

\begin{figure*}
\begin{center}
  \includegraphics[width=\textwidth]{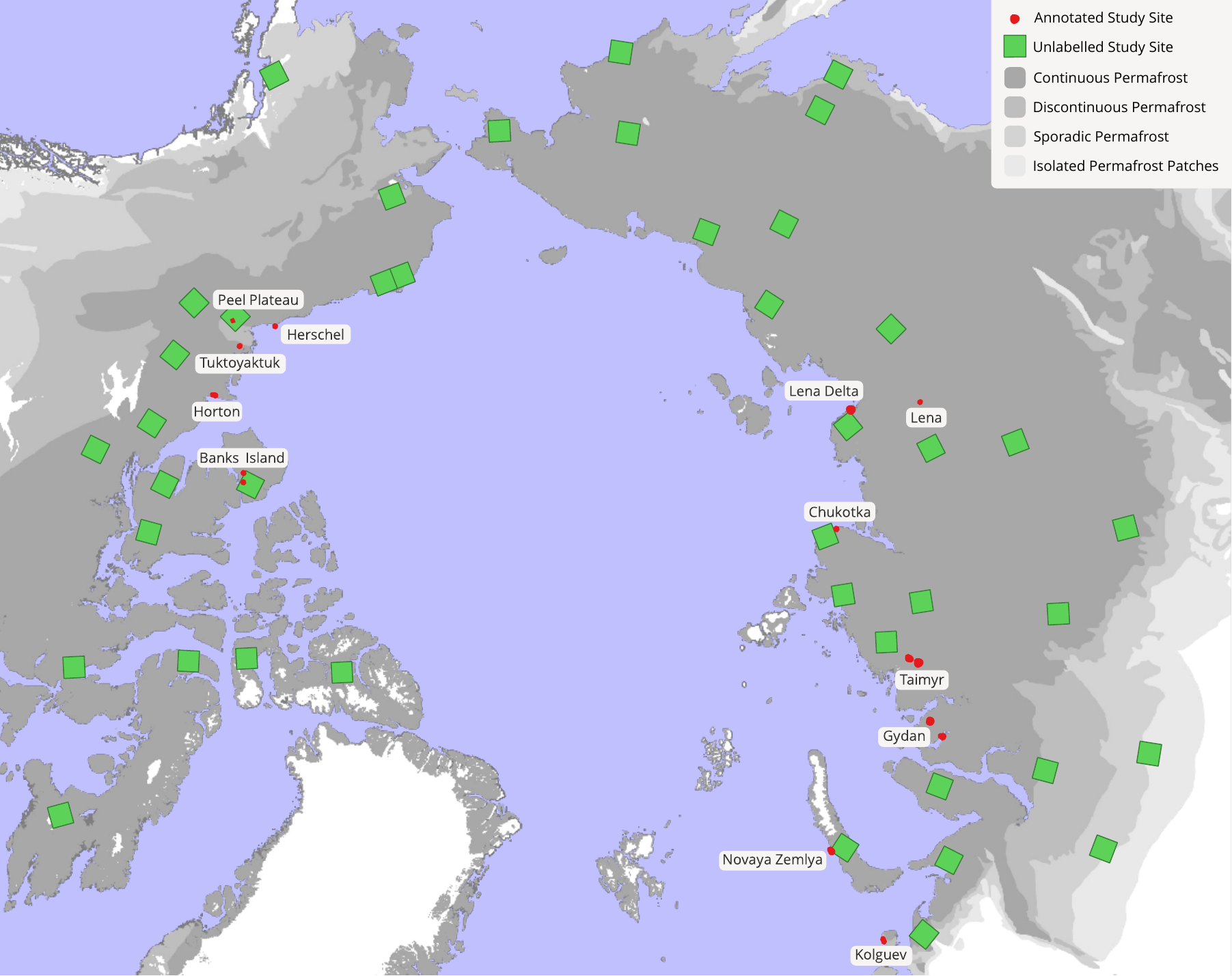}
\end{center}
\caption{Spatial distribution of the annotated training sites (red).
  It can be seen that the labelled data has quite limited spatial coverage.
  By using semi-supervised learning, it is possible to include large areas
  of unlabelled Sentinel-2 imagery (green) into the training process.
  Basemap source: \cite{brown2002_circumarctic}
}
\label{fig:dataset}
\end{figure*}

\section{Related Work}\label{sec:related}

\subsection{Monitoring Permafrost Disturbances}
As permafrost cannot be directly seen from space,
many remote-sensing studies for permafrost detection
focus instead on monitoring specific targets that are known or assumed to be correlated
with the state of permafrost or its vulnerability~\cite{bartsch2023_permafrost}.
Especially the spatially consistent and temporally high resolution 
monitoring of specific permafrost degradation landforms is a desirable
goal, since these also allow assessments
regarding vulnerability of local infrastructure
and the biogeochemical implications
of rapid permafrost thaw
for both the local environment and the global climate system.

Regarding data sources, permafrost disturbances
can be mapped using different remote sensing approaches,
like optical image analysis~\cite{nitze2021_developing},
optical time series analysis~\cite{brooker2014_mapping},
or interferometric
synthetic aperture radar (InSAR) measurements~\cite{bernhard2020_mapping}.

Many studies rely on manual digitization of permafrost disturbance landforms
in satellite imagery~\cite{segal2016_acceleration,leibman2023_distribution}.
While this approach ensures good accuracy, it quickly becomes infeasible
when the study areas grow beyond small to medium sized regions.
In order to automate the laborious manual digitzation process,
some studies explored computer vision methods like
trend analyses combined with random forests~\cite{nitze2018_remote},
or graph-based analysis~\cite{rettelbach2021_quantitative}.

With deep learning becoming an indispensable tool in remote sensing,
it was also used for the detection of RTS features.
Huang et al.~\cite{huang2018_automatic} adapted the DeepLab architecture
for semantic segmentation~\cite{chen2018_deeplab} to the task of
mapping permafrost features like RTS
using imagery from unmanned aerial vehicles (UAVs) over the
northeastern Tibetan Plateau.
Similarly, Nitze et al.~\cite{nitze2021_developing} trained several CNN architectures
on PlanetScope satellite imagery for 6 study sites in northwest Canada and
the Russian Arctic.

Existing studies usually focus on a single region of interest, like
the Canadian Arctic~\cite{huang2022_accuracy},
the Tibetan Plateau~\cite{huang2018_automatic,huang2020_using},
or a few select regions~\cite{nitze2018_remote,nitze2021_developing,yang2023_mapping}.

Apart from the previously introduced RTS, 
there are various other disturbances of permafrost
that can also be mapped and monitored using remote sensing techniques:
\begin{itemize}
  \item \emph{Thermokarst lakes} are formed by the thawing of ice-rich permafrost
    and can be mapped well from
    space~\cite{nitze2018_remote,hughes-allen2023_automated}.
    They are of great interest because of their role
    in the permafrost carbon feedback loop to climate.
  \item \emph{Wildfires} can greatly change the landcover,
    and are considered a major driver of permafrost thaw.
    Both active fires and burn scars can be observed in
    remotely sensed imagery~\cite{gibson2018_wildfire,nitze2018_remote}.
  \item While not directly visible from space, \emph{ice wedges} and their degradation
    can be mapped through the characteristic polygonal patterns they induce
    on the land surface in permafrost lowlands ~\cite{abolt2019_brief,rettelbach2021_quantitative,witharana2021_objectbased}.
\end{itemize}

\subsection{Semi-supervised Semantic Segmentation in Remote Sensing}
In remote sensing, many relevant tasks are semantic segmentation tasks.
For each pixel, a class label needs to be predicted in order to
partition the entire scene into separate regions of interest.
Such tasks are encountered across a large number of research areas like
crop type mapping~\cite{kondmann2021_denethor},
urban mapping~\cite{volpi2017_dense},
or monitoring animal populations~\cite{bowler2020_using}.
Generally, it is quite hard even for experts to perfectly
annotate a given scene pixel by pixel,
and the process of generating these annotations
is often tedious and time-consuming. 
There are approaches to reducing the labelling burden through working with sparse labels
like point labels or scribbled labels, but these come at a price in terms of classification accuracy~\cite{hua2022_semantic}.
On the other hand, unlabelled remote sensing data
is generally easily available through programmes like
NASA's Landsat series or ESA's Copernicus missions.
Therefore, the idea of combining small labelled datasets
with large unlabelled data for semantic segmentation has been previously
explored in remote sensing.

A large class of semi-supervised learning studies in remote sensing
focuses on the idea of consistency regularization.
The underlying assumption here is that even for unlabelled images,
a model's representations or outputs should be consistent under
a certain set of perturbations.
For example, these perturbations can be
data augmentation operations~\cite{upretee2022_fixmatchseg},
feature dropout~\cite{lee2013_pseudolabel},
additive noise in the feature space~\cite{li2022_semisupervised,yang2023_revisiting},
or interpolation between samples~\cite{verma2022_interpolation}.
Under these perturbations,
the model is then trained to stay consistent.
This consistency can be enforced at different
stages of the model calculation.
Most common is the so-called pseudo-labelling technique~\cite{lee2013_pseudolabel},
where consistency is enforced in the final output classification of the network.
Various extensions of this basic
idea exist~\cite{sohn2020_fixmatch,zhang2022_semisupervised}.
Another possibility is to enforce consistency in the intermediate feature space
within a given layer of the neural network~\cite{li2022_semisupervised}.
Such approaches have been successfully applied for
mapping building footprints~\cite{li2022_semisupervised},
mapping landslides~\cite{zhang2022_generalization}
or aerial image segmentation~\cite{wang2020_semisupervised}.
Our presented approach is similar to these methods.
The main difference in our approach is the change
from pseudo-labels to pseudo-classes.
While pseudo-labels are adhering to the original classification scheme of the task,
we allow the network to come up with additional classes in order to
oversegment the images.
This should be particularly helpful for tasks with a large class imbalance,
for example when a background class with high intraclass variance
dominates the scenery, which is the case in RTS detection.

The Generator-Discriminator approach from
Generative Adversarial Networks (GANs) has also been explored
for semi-supervised semantic segmentation.
Here, the basic idea is to conceptually understand the 
segmentation network as either the generator or the discriminator network.
In the first setup,
the discriminator learns to discern true segmentation maps from model outputs
on a pixel-wise level.
At the same time, the segmentation network takes the role of the generator
and is trained to fool the discriminator as a secondary
loss objective~\cite{hung2018_adversarial}.
In the other setting, a generator is used to generate fake data,
and the discriminator is trained to differentiate these fake data points
from the unlabelled data, while also generating class labels~\cite{souly2017_semi}.
Adversarial semi-supervised learning approaches have been demonstrated on tasks
like hyperspectral image classification~\cite{he2017_generative}
or change detection~\cite{liu2019_semisupervised}.
Other than these works,
our method only requires training a single neural network.
Also, it does not exhibit the well-known training instabilities or
require any of the careful hyperparameter tuning that adversarial methods are known for.

Finally, some studies separate the training process into
a self-supervised pre-training phase on a large unlabelled dataset,
and a supervised fine-tuning phase on the labelled dataset.
As self-supervised has been an area of great interest in computer vision
recently, this approach is getting increasingly popular.
For example, such approaches have been shown to improve model performance
for tasks such as hyperspectral image classification~\cite{braham2022_self},
land cover mapping~\cite{heidler2023_selfsupervised,manas2021_seasonal}
or change detection~\cite{manas2021_seasonal}.
Contrasting this, we present a semi-supervised training procedure
where the model is trained end-to-end in a single training phase.

\begin{figure*}
\begin{center}
  \includegraphics[width=\textwidth]{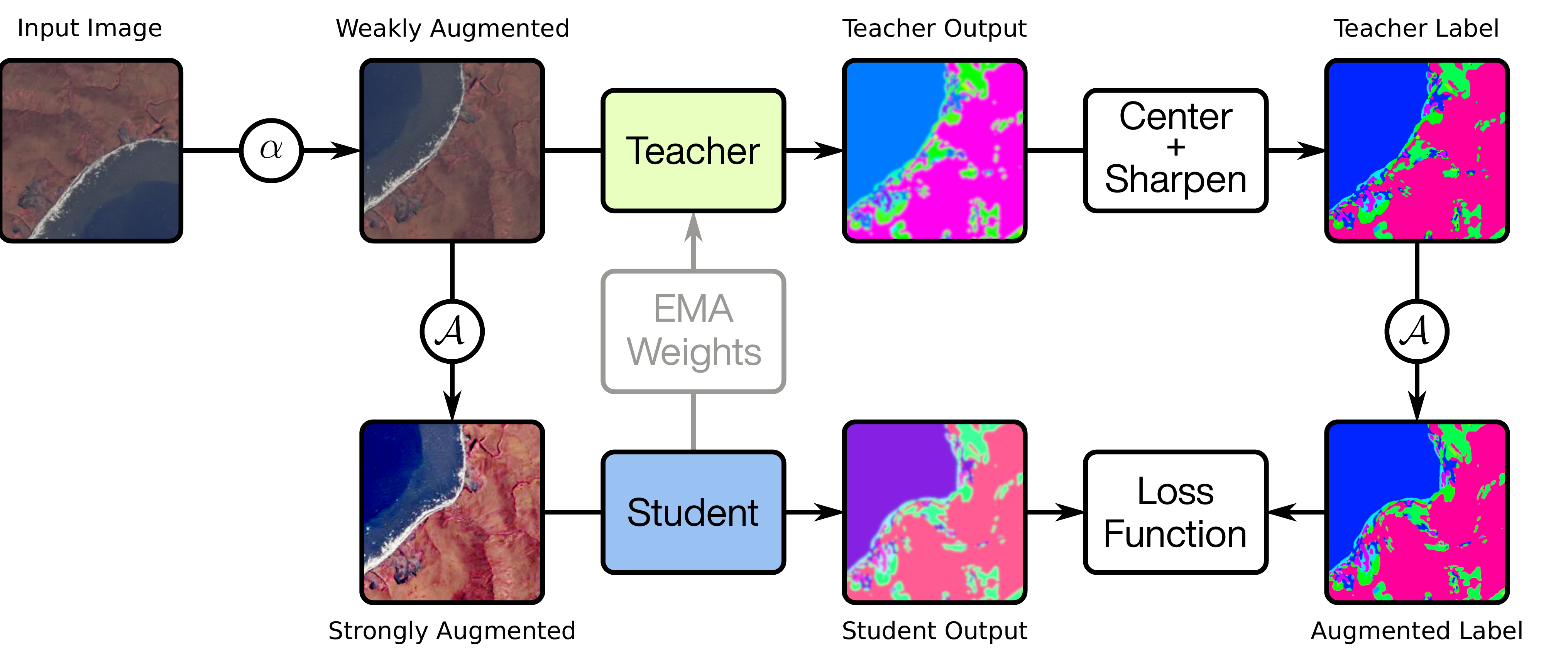}
\end{center}
\caption{%
  Overview of the self-supervised part of the
  PixelDINO framework for pixel-wise feature learning.
  First, the image is weakly augmented and a dense feature map is
  derived using the teacher model.
  These teachers are turned into class labels by centering, sharpening,
  and applying the softmax function.
  Both the weakly augmented image and the teacher label
  are augmented using the set of strong augmentations.
  The student model is then trained on this pair of image and label.
  Finally, the teacher model's weights are updated as an exponential
  moving average of the student's weights.
}\label{fig:architecture}
\end{figure*}

\section{PixelDINO for Semi-Supervised Semantic Segmentation}

The DINO framework~\cite{caron2021_emerging}
was originally developed for self-supervised pre-training of
image models like CNNs and Vision Transformers for image classification.
The main concept introduced in this framework
is the so-called \emph{self-distillation with no labels}.
In this process, the model is tasked with defining its own
classification scheme for images.
Two versions of the model, called student and teacher,
are trained following the self-distillation process.

For a given input image, two augmentations are generated.
Out of these two augmentations, the first one is run through the teacher model.
The features derived by the teacher model are then centered and re-scaled.
Finally, the teacher's classification is derived by applying a softmax
activation to the re-scaled outputs.
Meanwhile, the second version of the image is
run through the student model.
Finally, the student is then trained to match the teacher's classifications
with its own outputs~\cite{caron2021_emerging}.
In the following, we will be referring to the classes
automatically derived by the models
as ``pseudo-classes''.

Naturally, one crucial step in this setup is the assignment
of parameters to the teacher model.
As there are no ground-truth labels in this setup,
the teacher weights are taken to be an exponential moving average (EMA)
of the student weights, hence the term ``self-distillation''.
In this way, the student network is quickly evolving in different directions
of the parameter space
while at the same time being pulled towards consistency with the teacher model.
Meanwhile, the teacher network is slowly following the student network with its parameters.

\subsection{Self-supervised learning of pixel features}
While natural imagery often has a clear object of focus,
a remotely sensed satellite image can have
dozens or hundreds of objects of interest in it.
Therefore, we believe that working on the pixel level
will lead to more discriminative features,
which will be crucial for a successful segmentation of these objects in the end.
The basic idea for our PixelDINO framework is to adopt
the aforementioned DINO training process on a pixel-wise level.
Instead of classifying entire images, the student and teacher models will instead
give a label to each pixel in the input image.

But in the original DINO framework,
the labels generated by the teacher
can be directly applied to train the student.
A bit more care needs to be taken for the pixel-wise case.
Data augmentations like flips or rotations will change the location of objects
in the image.
Therefore, the pixel-wise segmentation labels also need to be augmented in the
same fashion.
When following the original DINO setup, doing this correctly is challenging,
as it would involve inverting the data augmentations applied to the first image.
Further, this procedure will introduce invalid pixel labels
when inverting lossy augmentations like
rotations by non-multiples of 90° or cropping operators.
To avoid these issues, we resort to an approach introduced
by the FixMatch~\cite{sohn2020_fixmatch,upretee2022_fixmatchseg}
line of semi-supervised learning.
Instead of using two augmentations of the same base image,
we will use a chain of augmented images.

With FixMatch, Sohn et al.~\cite{sohn2020_fixmatch}
developed a framework for semi-supervised learning.
One core idea behind this framework is to
use two sets of data augmentatations called
\emph{weak augmentations}, denoted by $\alpha(\cdot)$,
and \emph{strong augmentations}, denoted by $\mathcal A(\cdot)$.
Upon trying to generalize this scheme for semantic segmentation,
Upretee and Khanal~\cite{upretee2022_fixmatchseg} were facing the same issues
when transferring labels between two differently augmented versions of the same image.
As the labels themselves are also subject to geometric transformations
such as rotations,
converting them between augmentations gets
more involved when dealing with dense predictions such as
semantic segmentation.
Their proposed solution is to
chain the weak and strong data augmentations as $\mathcal A(\alpha(\cdot))$,
so that the pseudo-label can be augmented alongside with
the image~\cite{upretee2022_fixmatchseg}.
We use this idea to enable training of pixel-wise features in our PixelDINO
framework.

So given an unlabelled input image $U$,
we first apply a weak augmentation $\alpha(U)$ and
calculate the teacher output $\mathcal{T}(\alpha(U))$.
Then, the teacher's label is derived through centering,
re-scaling, applying the softmax function, and finally applying a strong augmentation:
\begin{equation}
  Y_U = \mathcal{A}\left(\operatorname{softmax}\left(
    \frac{\mathcal T(\alpha(U)) - \mu}
    \tau
  \right)\right)\\
\end{equation}
Here, $\mu$ is the center of past teacher outputs, which is updated using an
exponential moving average,
and $\tau$ is the temperature parameter.
A lower temperature leads to a stronger ``sharpening'' of the
class distribution,
which is desired in order to bias the model away from outputting
a uniform distribution.

The student model $\mathcal S$ is applied to the strongly augmented
input image to obtain the student's prediction
$\mathcal{S}(\mathcal{A}(\alpha(U)))$.
Finally, the PixelDINO loss is
calculated as the cross entropy between the softmax of the student output and the
teacher label:
\begin{equation}
  \mathcal{L}_\text{PixelDINO} = \operatorname{CrossEntropy}(%
  \operatorname{softmax}(\mathcal{S}(\mathcal{A}(\alpha(U))),
  Y_U)
\end{equation}

In this way, the student model $\mathcal S$ is trained to align
its predictions in such a way that they are consistent
with the teacher's outputs $\mathcal T$ under the set of
strong augmentations $\mathcal A$.
A graphical overview of this approach is given in Fig.~\ref{fig:architecture}.

\subsection{Semi-supervised learning with PixelDINO}

The goal for semi-supervised learning is to exploit the information
present in a large, unlabelled dataset and combine that with the class information
from a smaller, labelled dataset.
For PixelDINO, embedding the information from a labelled dataset
is rather straight-forward.
The DINO methodology already works with pseudo-classes,
and PixelDINO extends that to pseudo-classes per pixel.
If information about some specific classes is already known a priori
in the form of a labelled dataset,
this can be embedded into the training process in order
to make the pseudo-classes align with the a priori classes.
In our case, we would like to do exactly that
for the RTS class from the labelled dataset.

To achieve that,
we combine the PixelDINO training loop with a regular supervised training loop.
This means that during each training step, the student model will be trained
on both a mini-batch of labelled examples,
as well as one of unlabelled examples.
For a labelled example given as a pair of
an image $X \in \mathbb{R}^{C \times H \times W}$
and a mask $Y \in \mathbb{R}^{H \times W}$,
the supervised loss term is the regular cross-entropy which is commonly used
in semantic segmentation.
In practice, we also apply weak and strong data augmentation to the labelled
samples:
\begin{equation}
  \mathcal{L}_\text{supervised}(X, Y) = \operatorname{CrossEntropy}(%
    \mathcal{S}(\mathcal{A}(\alpha(X))), Y
  )
\end{equation}

The final, semi-supervised training objective is simply the weighted
sum of the two loss terms, balanced by a hyper-parameter $\beta$:
\begin{equation}
  \mathcal{L}_\text{semi-supervised}(X, Y, U) =
  \mathcal{L}_\text{supervised}(X, Y) +
  \beta \mathcal{L}_\text{PixelDINO}(U)
\end{equation}
Empirically, we set $\beta = 0.1$ for our experiments.

\begin{algorithm}[t]
\caption{Semi-supervised PixelDINO (Pytorch-style)}\label{alg:pixeldino}
\definecolor{codeblue}{rgb}{0.25,0.5,0.5}
\lstset{
  backgroundcolor=\color{white},
  basicstyle=\fontsize{9pt}{9pt}\ttfamily\selectfont,
  columns=fullflexible,
  breaklines=true,
  captionpos=b,
  commentstyle=\fontsize{9pt}{9pt}\color{codeblue},
  keywordstyle=\fontsize{9pt}{9pt}\bfseries,
}

\vspace{1mm}
\small
Hyper-Parameters:\\
\phantom{nd}\texttt{beta}: Weight of DINO loss\\
\phantom{nd}\texttt{temp}: Temperature used for softmax-scaling
\begin{lstlisting}[language=python]
def train_step(img, mask, unlabelled):
  # Supervised Training Step
  pred = student(img)
  loss_supervised = cross_entropy(pred, mask)

  # Get pseudo-classes from teacher
  view_1 = augment_weak(unlabelled)
  mask_1 = teacher(mask_1)
  mask_1 = (mask_1 - center) / temp
  batch_center = center.mean(dim=[0,2,3])
  mask_1 = softmax(mask_1)

  # Strongly augment image and label together
  view_2, mask_2 = augment(view_1, mask_1)

  pred_2 = student(view_2)
  loss_dino = cross_entropy(pred_2, mask_2)

  loss = loss_supervised + beta*loss_dino
  loss.backward()  # Back-propagate losses
  update(student)  # Adam weight update
  ema_update(teacher, student)  # Teacher EMA
  ema_update(center, batch_center)  # Center EMA
\end{lstlisting}
\end{algorithm}

The pseudo-code for this training procedure is outlined in Alg.~\ref{alg:pixeldino}.
By forcing the student model to adhere to the teacher outputs and the labelled
ground truth masks at the same time, it is very likely that the classification
schemes will indeed align to include one class for our desired target.

\subsection{Data Augmentations}\label{sec:aug}
Data Augmentation is a commonly used technique to make models more robust
to perturbations in the input, as well as encourage equivariance under
certain geometric transformations like rotations or reflections.
Further, it is a crucial component for semi-supervised learning,
which is why we will briefly explain the employed data augmentation techniques.

The semi-supervised learning methods introduced in this study
require two different sets of data augmentation operations,
in order to generate different views of the same data.
Following the terminology of Sohn et al.~\cite{sohn2020_fixmatch},
we separate the augmentations used in our study into \emph{weak augmentations},
denoted by $\alpha(\cdot)$,
and \emph{strong augmentations}, denoted by $\mathcal{A}(\cdot)$.
The conceptual difference is that weak augmentations should only add variation to the data
without making the classification more difficult.
Strong augmentations, on the other hand,
distort the image in such a way that makes it harder for the
model to perform the classification.

\subsubsection{Weak Augmentations}\label{sec:weak_aug}
In the class of weak augmentations,
we only include the simple geometric transformations
introduced before, namely horizontal and vertical reflections of the input imagery,
as well as rotations by multiples of 90°.
These augmentations are very frequently used in remote sensing as
models are expected to be equivariant under reflections and rotations
for many tasks.

\subsubsection{Strong Augmentations}\label{sec:strong_aug}
Designing a class of strong augmentations for remote sensing imagery is considerably harder
than weak augmentations.
The commonly used colorspace transformations which are often used for RGB imagery
do not generalize well to multi-spectral imagery.
Therefore, we settle for two classes of adjustments.
The first one consists of operations that change the image brightness and contrast,
namely the \texttt{RandomBrightness}, \texttt{RandomGamma}, and \texttt{RandomContrast} transformations.
The second group constsists of distorting geometrical transformations,
namely rotating by arbitrary angles (\texttt{Rotate}),
elastic transforms (\texttt{Warp}) and \texttt{Gaussian Blur}.

\section{Experiments \& Results}

\subsection{Data Acquisition and Preprocessing}\label{sec:dataset}
As the main data source for this study,
we use the openly available RTS inventory from Nitze et al.~\cite{nitze2021_developing}%
\footnote{available at \url{https://github.com/initze/ML_training_labels}}.
This inventory consists of polygons that were manually labelled using PlanetScope imagery,
elevation data, and Landsat timeseries as the source data.
Its extent amounts to 4335 polygon annotations of
RTS footprints from the years of 2018 and 2019,
with a combined area of \SI{\sim 84}{\kilo\metre\squared}.
The focus of the inventory lies on multiple regions in the continental Arctic,
mostly in coastal areas.

\subsection{Generalization Study}\label{sec:generalization_study}
\begin{table*}
  \caption{Results of the Generalization Study:
  Mean and Standard Deviation of 4 runs each (Values in \%)}\label{tab:results}
  \center\begin{tabular}{lcccccccc}
\toprule
 & \multicolumn{4}{c}{Herschel} & \multicolumn{4}{c}{Lena} \\
 & IoU & F1 & Precision & Recall & IoU & F1 & Precision & Recall \\
 \cmidrule(lr){2-5} \cmidrule(l){6-9}

Baseline & 19.8 \textcolor{gray}{± \phantom{0}1.7} & 33.0 \textcolor{gray}{± \phantom{0}2.3} & 28.8 \textcolor{gray}{± \phantom{0}3.0} & 39.4 \textcolor{gray}{± \phantom{0}5.0} & 28.8 \textcolor{gray}{± \phantom{0}4.0} & 44.6 \textcolor{gray}{± \phantom{0}5.0} & 52.8 \textcolor{gray}{± \phantom{0}5.9} & 39.0 \textcolor{gray}{± \phantom{0}6.0} \\
Baseline+Aug & 22.9 \textcolor{gray}{± \phantom{0}3.0} & 37.2 \textcolor{gray}{± \phantom{0}3.9} & 44.2 \textcolor{gray}{± \phantom{0}7.5} & 32.3 \textcolor{gray}{± \phantom{0}2.0} & 25.8 \textcolor{gray}{± 10.2} & 40.2 \textcolor{gray}{± 13.0} & 69.4 \textcolor{gray}{± \phantom{0}3.2} & 29.4 \textcolor{gray}{± 12.5} \\
\midrule
FixMatchSeg~\cite{upretee2022_fixmatchseg} & 23.4 \textcolor{gray}{± \phantom{0}0.8} & 37.9 \textcolor{gray}{± \phantom{0}1.1} & 34.1 \textcolor{gray}{± \phantom{0}2.3} & 43.2 \textcolor{gray}{± \phantom{0}4.5} & 32.4 \textcolor{gray}{± \phantom{0}3.2} & 48.8 \textcolor{gray}{± \phantom{0}3.7} & 59.4 \textcolor{gray}{± \phantom{0}2.7} & 41.6 \textcolor{gray}{± \phantom{0}5.0} \\
Adversarial~\cite{hung2018_adversarial} & 26.6 \textcolor{gray}{± \phantom{0}3.9} & 41.9 \textcolor{gray}{± \phantom{0}4.9} & 60.0 \textcolor{gray}{± \phantom{0}9.2} & 32.3 \textcolor{gray}{± \phantom{0}3.1} & 25.1 \textcolor{gray}{± 15.1} & 38.2 \textcolor{gray}{± 20.5} & 87.3 \textcolor{gray}{± \phantom{0}7.5} & 26.8 \textcolor{gray}{± 16.7} \\
PixelDINO & 30.2 \textcolor{gray}{± \phantom{0}2.7} & 46.4 \textcolor{gray}{± \phantom{0}3.2} & 52.7 \textcolor{gray}{± \phantom{0}9.2} & 42.0 \textcolor{gray}{± \phantom{0}3.0} & 39.5 \textcolor{gray}{± \phantom{0}6.5} & 56.4 \textcolor{gray}{± \phantom{0}6.6} & 77.7 \textcolor{gray}{± \phantom{0}6.3} & 44.5 \textcolor{gray}{± \phantom{0}6.8} \\
\bottomrule
\end{tabular}

\end{table*}

While Nitze et al.~\cite{nitze2021_developing} base their analyses on PlanetScope imagery,
we opt for Sentinel-2 imagery for this study due to its open availability,
which is an important factor for building a large unlabelled
dataset for semi-supervised learning.
Practically speaking, these two satellite platforms mainly differ in their imaging resolution and
their spectral channels.
While PlanetScope imagery is provided at ground sampling distances of \SIrange{3}{4}{\metre}
and contains the visible RGB channels as well as a near-infrared channel,
Sentinel-2 imagery comes at a reduced spatial resolution of \SI{10}{\metre} per pixel,
but in turn features 13 spectral channels.

Using the image\href{}{} footprints from the RTS inventory,
we next download 83 matching Sentinel-2 Level 1C imagery sourced from Google Earth Engine. 
As the last step, the RTS annotation polygons are rastered to match the satellite image pixel grids.
The annotation masks then contain the binary values 0 and 1
for background and RTS pixels, respectively.

For the unlabelled semi-supervised training dataset,
we arbitrarily select 42 Sentinel-2 tiles
over permafrost areas with a focus on
regions of continuous permafrost with high estimated ice content.
For each one of these tiles,
we then randomly select a year from the Sentinel-2 acquisition range and
download the least cloudy tile taken between May and August of that year.

The obtained Sentinel-2 scenes are much larger than even modern
GPU cards can handle for neural network training.
Further, mini-batch training requires a uniform image size.
To fulfill these requirements,
all imagery is cut into patches of size $192 \times 192$ pixels
as part of the training pipeline.

After all pre-processing steps,
we arrive at a labelled training dataset with 6464 patches,
an unlabelled training dataset with 266168 patches,
and two test datasets, Herschel and Lena,
with 1052 and 4420 patches, respectively.

\subsection{Training Details}
For each configuration,
we train 4 models with different random seeds
to also quantify the effects of the randomness in model initialization,
mini-batch sampling, and data augmentation.
Models were trained on a GPU server equipped with NVIDIA A6000 GPUs.
The implementation was carried out in JAX~\cite{bradbury2018_jax} and
Haiku~\cite{hennigan2020_haiku}.
The code is available online at \url{https://github.com/khdlr/PixelDINO}.

In the semi-supervised setting,
the model is being trained on two datasets,
the labelled data and the unlabelled data.
These two datsets are vastly different in size,
with the labelled dataset being much smaller than the
unlabelled dataset.
Therefore, the concept of ``training epochs'' is no longer appropriate
for specifying the training duration of the model.
In order to still keep comparable training schedules
for the different model configurations,
we instead count the number of training steps applied to each model.
This should keep the comparison between the models as fair as possible,
as each model has gone through the same training schedule.
In all reported experiments, the models were trained for 200\,000 steps.

\begin{figure*}
  \begin{center}
  \setlength\tabcolsep{2pt}
  \begin{tabular}{cc}
    \textbf{Supervised}&\textbf{PixelDINO}\\
    \includegraphics[width=0.4\linewidth]{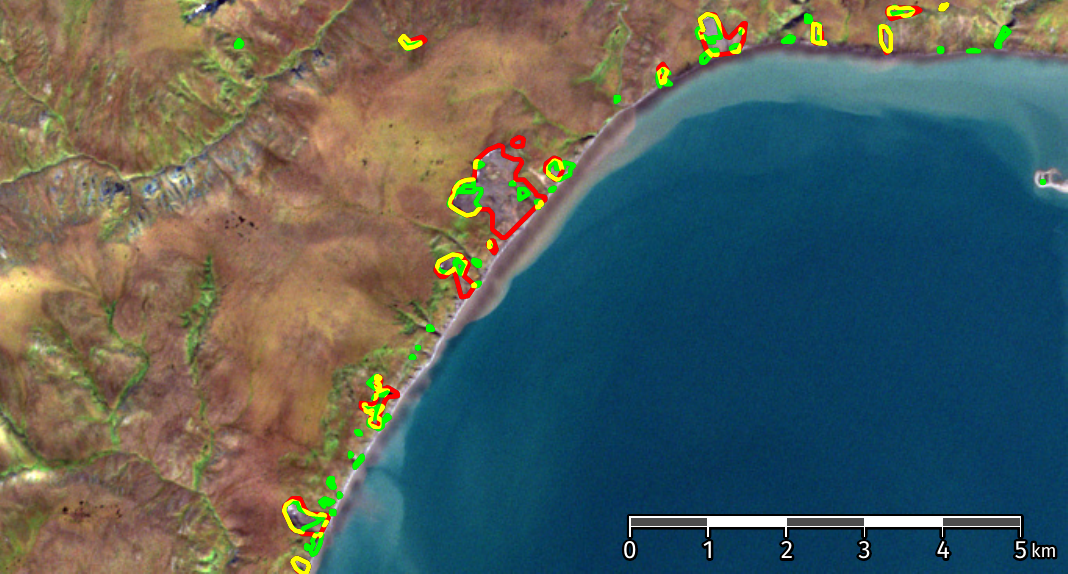}&
    \includegraphics[width=0.4\linewidth]{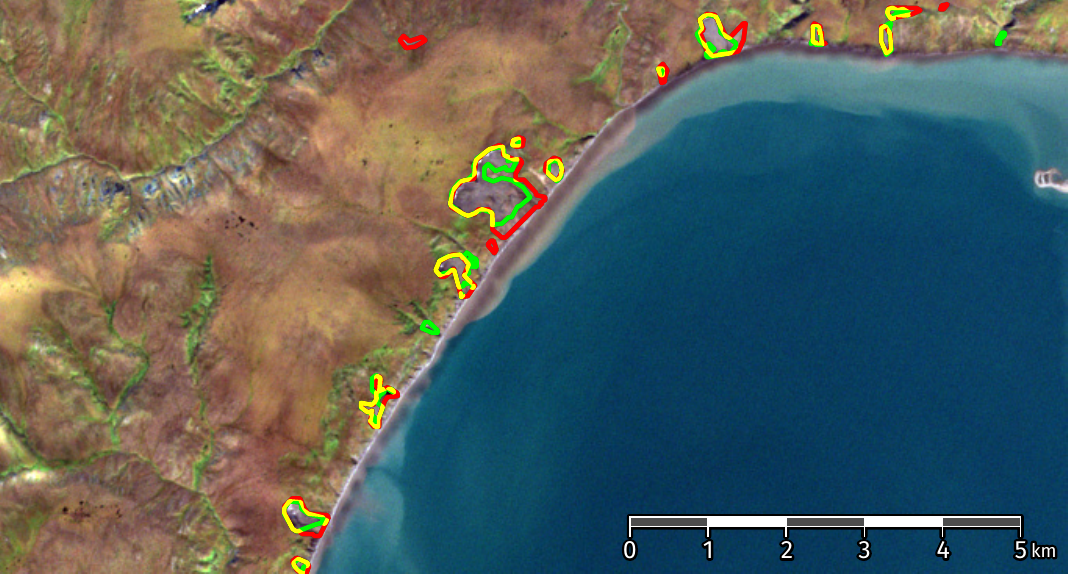}\\
    \includegraphics[width=0.4\linewidth]{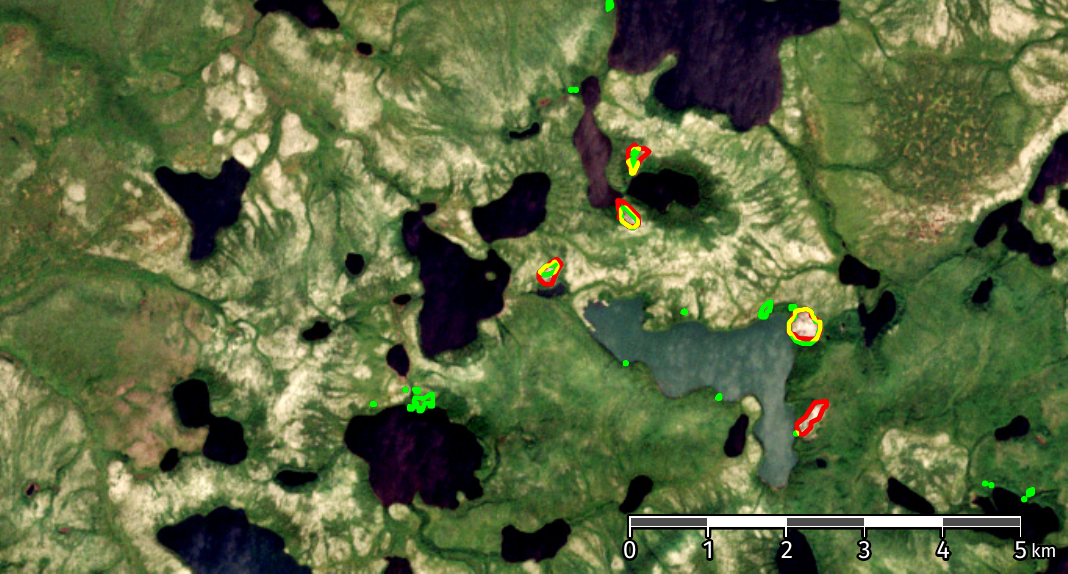}&
    \includegraphics[width=0.4\linewidth]{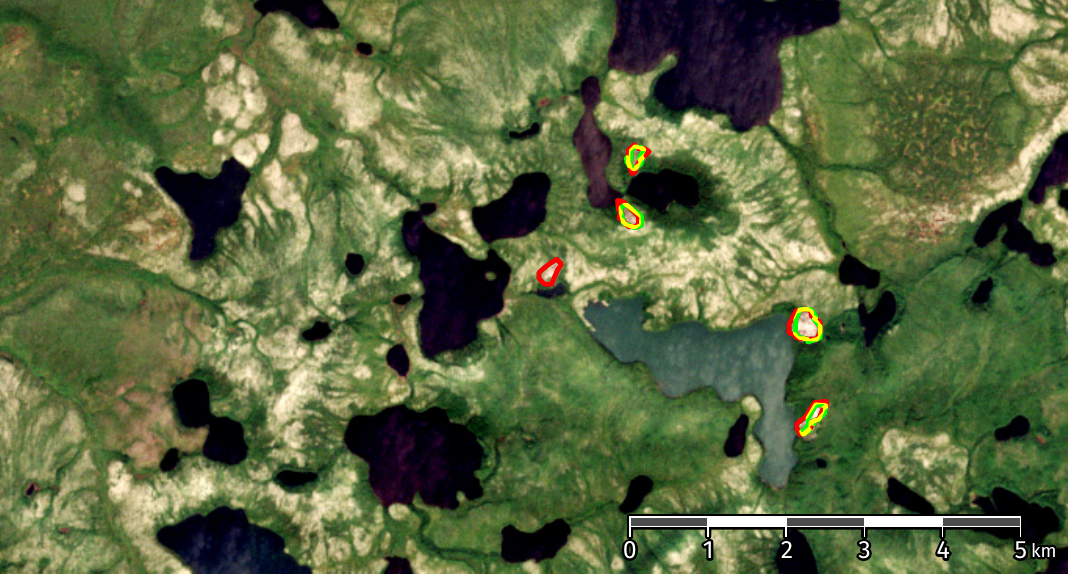}\\
  \end{tabular}\\
  $\vcenter{\hbox{\tikz{\path[draw=red,line width=0.5mm] (0,0) -- (6mm,0);}}}$ Ground Truth\hspace{2mm}
  $\vcenter{\hbox{\tikz{\path[draw=green,line width=0.5mm] (0,0) -- (6mm,0);}}}$ Prediction\hspace{2mm}
  $\vcenter{\hbox{\tikz{\path[draw=yellow,line width=0.5mm] (0,0) -- (6mm,0);}}}$ Overlapping GT and Prediction
  \end{center}
  \caption{
    Prediction results for parts of the Herschel Island (top) and Lena (bottom)
    study sites for the Baseline+Aug and PixelDINO training methods.
    Most prominent is the large reduction in false positives
    due to the semi-supervised training method.
  }\label{fig:results}
\end{figure*}

Out of the annotated study regions in the original dataset,
we set aside the Herschel Island and Lena sites for testing purposes.
We chose the Herschel Island site for being spatially separated from the
Canadian mainland.
On the other hand, the Lena test site was chosen because it is the only
study site that is not situated on the coast of the Arctic Ocean.
Instead, it is located further upstream the Lena River.
As the models will not encounter labelled examples for regions this far
inland, the evaluation results on this test site will be especially
interesting for quantifying the generalization capabilities of the
models to completely new regions.
All of the remaining annotated regions are used as the labelled training set.
Additionally, the unlabelled Sentinel-2 tiles are used during training
for the semi-supervised approaches (cf.~Fig.~\ref{fig:dataset}).

In order to quantify the improvements from the modified training procedure,
we conduct experiments with different configurations.
Starting with a baseline study without any training improvements,
we keep the model architecture constant while
making modifications to the training process.
For good comparability, we also use both the weak and strong data augmentations
we defined in section~\ref{sec:aug} for this experiment.

Specifically, we train and evaluate models in the following configurations:

\subsubsection{Baseline}
  Models trained only using supervised learning,
  without any data augmentation,
\subsubsection{Baseline+Aug}
  Same as baseline, but trained using the weak and strong
  data augmentation as described
  in section~\ref{sec:aug}.
\subsubsection{FixMatchSeg}
  Models trained in the semi-supervised setting
  using the methodology described by Upretee and Khanal~\cite{upretee2022_fixmatchseg}.
\subsubsection{Adversarial}
  Semi-supervised models trained using the adversarial approach
  proposed by Hung et al.~\cite{hung2018_adversarial}.
\subsubsection{PixelDINO}
  Models trained in the semi-supervised setting
  using our proposed methodology
  as outlined in Alg.~\ref{alg:pixeldino}.

As the introduced methodology focuses on adapting the training process itself
rather than making changes to the model architecture,
it is invariant to the specific model architecture used.
Therefore, any semantic segmentation model can be used in practice.
For our experiments,
we use the UNet model~\cite{ronneberger2015_unet} as it is a widely used
network architecture for image segmentation tasks in remote sensing.

The foreground and background classes in this dataset are highly imbalanced.
Even though the study areas were chosen to feature regions of high RTS density,
only around 0.7\% of all pixels contain a target, while all other pixels
belong to the background class.
Therefore, pixel-wise accuracy is an unfit metric for this task.
Instead, we evaluate the models using other metrics which are widely used for
such imbalanced segmentation tasks:
\begin{enumerate}
  \item Intersection over Union (IoU): Fraction of true positives pixels among all pixels
    that are true targets and/or classified positive.
  \item Precision: Fraction of true positive pixels among positive classifications.
  \item Recall: Fraction of true positive pixels among true target pixels.
  \item F1 score: The harmonic mean of Precision and Recall.
\end{enumerate}

The evaluation results of the generalization study are displayed in
Tab.~\ref{tab:results}.
Overall, the trend shows better performance of semi-supervised learning methods
compared to the supervised baselines.
Among the semi-supervised methods, our proposed PixelDINO approach
demonstrates the strongest performance.

\section{Discussion}

The results show that for the task of RTS detection,
semi-supervised learning can indeed yield a strong performance boost.
In this section we will discuss our observations during the experiments,
what sets apart PixelDINO from the other semi-supervised learning methods,
and implications for follow-up research.

\subsection{Isolating the Effect of Data Augmentations}
As consistency across data augmentations makes up a
large part of the semi-supervised training methods,
the improvements in segmentation accuracy might in fact be explained by the use of
data augmentations instead of the semi-supervised training itself.
In order to isolate the direct effects of data augmentation on the training
process, we trained the baseline supervised model with and without data augmentations.

Surprisingly, the data augmentations improve the model performance
on the Herschel evaluation site,
but actually decreases performance for the
Lena evaluation site.
We attribute this to the fact that the Lena site is much more
geophysically different from the training sites than the Herschel site.
Therefore, simple data augmentation allows
the model to better detect coastal thaw slumps,
while the generalization performance to inland regions suffers slightly.

At the same time, semi-supervised learning improves the performance of the baseline
model much more than just applying data augmentations.
From this,
we conclude that the improved training performance is not explained
by the data augmentations alone,
but can instead be attributed to the semi-supervised learning methods.

\subsection{Benefits of Semi-Supervised Learning}
The results in Tab.~\ref{tab:results} show that
the evaluated semi-supervised methods were generally
able to improve over the baselines in terms of
the IoU and F1 metrics.
Overall, semi-supervised learning has a large positive influence
on the performance of the models,
with the potential to increase IoU scores
by around 8 basis points and F1 scores by around 12 basis points across both datasets.

The only exception here is the performance of the adversarially trained models
on the Lena evaluation site.
Here, this class of models actually underperforms the baselines on average.
At the same time, the standard deviation is quite high,
implying a large spread in model performances for this particular group.
This behavior is likely tied to the most common point of criticism for
adversarial training,
namely that the training objective dictates a saddle-point optimization problem.
These are known to be hard to solve and lead to
unstable training~\cite{saxena2021_generative}.
Meanwhile, FixMatchSeg and PixelDINO do not exhibit this issue.

Generally, our proposed PixelDINO methodology achieves the strongest improvement in the
segmentation metrics.
This confirms that it is not only competitive with
other approaches for semi-supervised semantic segmentation,
but, at least for this task, is in fact the preferrable option.

\subsection{Effects of PixelDINO Training}\label{sec:pixeldino_benefits}
Our hypothesis for the strong performance of PixelDINO models lies in the fact
that RTS detection is a task that has only two classes and a strong class imbalance.
Therefore, the consistency regularization in approaches based on pseudo-labels
like FixMatchSeg does not regularize the model sufficiently when it comes to
correctly segmenting background features.
This hypothesis is supported by visual inspection (see Fig.~\ref{fig:results})
and the Recall and Precision metrics in the Tab.~\ref{tab:results}.
While FixMatchSeg and PixelDINO have comparable Recall values,
PixelDINO is far ahead in Precision, which suggests that
our method is able to greatly reduce the number of false positives
while maintining a constant number of false negatives.

Interestingly, an inverted phenomenon can be observed for the adversarial training method.
Here, the Precision values are greatly increased, beating even the models trained with PixelDINO.
But this comes at the cost of poor Recall values, which means that the adversarially
trained model will miss many more RTS targets than the other methods.
We believe this to be related to the adversarial training method.
As the discriminator is tasked with discerning true masks from predicted masks,
it teaches the segmentation network mainly about the shapes
of the features.
While it is hard for the model to fake believable RTS shapes,
it is really easy to fake a believable background by not predicting any target.
So when in doubt, the adversarial model would rather not predict anything
than running the risk of predicting a wrong shape.

Overall, our PixelDINO approach greatly benefits from its ability
to further subdivide the background class
into regions of different semantic content, which makes the
semi-supervised training feedback much more valuable,
which in turn leads to more accurate predictions on the test set.

\subsection{Limitations}
One common concern with increasing the complexity of training schemes
is the amount by which they will increase training time.
In order to be transparent about this limitation, we report the
average runtime of our experiments from section~\ref{sec:generalization_study}
in Tab.~\ref{tab:runtime}.
While the impact of data augmentations on the training duration is negligible,
the semi-supervised training methods increase the duration of training
by a factor of around 2.
This is easily explained by the fact that the semi-supervised methods
process both a batch of labelled imagery and a batch of unlabelled imagery
during each iteration.
However, we stress that these duration increases only occur during training and not during inference.
During Inference, all the presented models will run at the same speed since they share the same
model architecture.

As described previously in section~\ref{sec:pixeldino_benefits},
we believe that dealing with highly imbalanced classes is a strong property of PixelDINO.
While our introduced framework is flexible in terms of the number of output channels,
further research is needed to understand how well PixelDINO will generalize
to semantic segmentation problems with many classes.

\begin{table}
  \caption{Runtime of the Evaluated Training Methods}\label{tab:runtime}
  \center
  \begin{tabular}{lrrr}
\toprule
Method & Training Duration & Change \\
\midrule
Baseline & 88.9 min & -- \\
Baseline+Aug & 91.3 min & + \phantom{00}2.7\% \\
\midrule
FixMatchSeg & 178.1 min & + 100.3\% \\
Adversarial & 182.4 min & + 105.2\% \\
PixelDINO & 174.9 min & + \phantom{0}96.8\% \\
\bottomrule
\end{tabular}

\end{table}

\section{Conclusion}
Large volumes of remote sensing data are readily available
to the public through platforms like the NASA Landsat or ESA Copernicus archives.
These open up many possibile use cases for monitoring applications.
The most difficult part of such projects is then often the data annotation process,
because it is such a time-consuming process.
This is particularly true for semantic segmentation tasks,
because these require all pixels to be labelled.
Semi-supervised learning can help relieve the labelling workload on domain experts
by a large amount, simply by using readily available unlabelled data.

Our proposed PixelDINO framework achieves this by encouraging the trained
model to come up with its own scheme of segmentation classes,
for which it is then trained to be consistent across data augmentations
as well as to align its classes to the label classes from the annotated
training set.

In our experiments we demonstrated that
PixelDINO can train models that generalize well
to previously unseen regions in the Arctic,
and do so better than both supervised baselines
and other semi-supervised approaches.

It is hypothesized that satellite imagery of higher resolution
will be beneficial for detecting RTS,
as oftentimes the targets can be quite small~\cite{nitze2021_developing}.
While we do not make use of such imagery due to reasons
of data availability,
the introduced methodology is applicable to any imagery source.
It is up to future research to explore the possibilities
of such methods for high-resolution satellite or even aerial imagery sources.

We believe the methods developed in this study are transferrable
to many different usecases in remote sensing even outside of permafrost monitoring.
Therefore we hope to inspire follow-up research in improving the automated mapping
of ground features using semi-supervised semantic segmentation methods.

\bibliographystyle{IEEEtran}
\bibliography{IEEEfull,references}

\end{document}